\documentclass{article}

\usepackage[preprint]{neurips_2024}

\usepackage{natbib}
\setcitestyle{numbers,square}

\usepackage[utf8]{inputenc} 
\usepackage[T1]{fontenc}    
\usepackage{hyperref}       
\usepackage{url}            
\usepackage{booktabs}       
\usepackage{amsfonts}       
\usepackage{nicefrac}       
\usepackage{microtype}      
\usepackage{xcolor}         

\usepackage{microtype}
\usepackage{graphicx}
\usepackage{subfigure}
\usepackage{booktabs} 

\usepackage{amsmath}
\usepackage{amssymb}
\usepackage{mathtools}
\usepackage{amsthm}

\usepackage{enumitem}
\usepackage{booktabs}
\usepackage{multirow}
\usepackage{multicol}
\usepackage{makecell}
\usepackage{array}
\usepackage{comment}
\usepackage{caption}

\usepackage{wrapfig}

\definecolor{brightpink}{rgb}{1.0, 0.0, 0.5}
\definecolor{ao(english)}{rgb}{0.0, 0.5, 0.0}
\definecolor{blue(ncs)}{rgb}{0.0, 0.53, 0.74}
\newcommand{\myPink}[1]{\textcolor{brightpink}{#1}}
\newcommand{\myRed}[1]{\textcolor{red}{#1}} 
\newcommand{\myGray}[1]{\textcolor{gray}{#1}} 
\newcommand{\myGreen}[1]{\textcolor{ao(english)}{#1}}
\newcommand{\myBlue}[1]{\textcolor{blue(ncs)}{#1}}

\title{Towards Weakly Supervised End-to-end Learning \\ for Long-video Action Recognition}

\author{Jiaming Zhou\textsuperscript{1},
Hanjun Li\textsuperscript{2},
Kun-Yu Lin\textsuperscript{3},
Junwei Liang\textsuperscript{1,4}\footnotemark[2]
\\
\textsuperscript{1}{AI Thrust, The Hong Kong University of Science and Technology (Guangzhou)}\\
\textsuperscript{2}{Tencent Youtu Lab} \ \ \ 
\textsuperscript{3}{Sun Yat-sen University}\\
\textsuperscript{4}{CSE, The Hong Kong University of Science and Technology}\\
\tt\small jzhou760@connect.hkust-gz.edu.cn junweiliang@hkust-gz.edu.cn
}

\begin{document}

\footnotetext[2]{Corresponding author}

\maketitle

\begin{abstract}
Developing end-to-end action recognition models on long videos is fundamental and crucial for long-video action understanding. 
Due to the unaffordable cost of end-to-end training on the whole long videos, existing works generally train models on short clips trimmed from long videos.
However, this ``trimming-then-training'' practice requires action interval annotations for clip-level supervision, i.e., knowing which actions are trimmed into the clips. Unfortunately, collecting such annotations is very expensive and prevents model training at scale.
To this end, this work aims to build a weakly supervised end-to-end framework for training recognition models on long videos, with only video-level action category labels. 
Without knowing the precise temporal locations of actions in long videos, our proposed weakly supervised framework, namely \textbf{AdaptFocus}, estimates where and how likely the actions will occur to \textbf{adapt}ively \textbf{focus} on informative action clips for end-to-end training.
The effectiveness of the proposed AdaptFocus framework is demonstrated on three long-video datasets.
Furthermore, for downstream long-video tasks, our AdaptFocus framework provides a weakly supervised feature extraction pipeline for extracting more robust long-video features, such that the state-of-the-art methods on downstream tasks are significantly advanced.
We will release the code and models.
\end{abstract}

\section{Introduction}
\label{sec:intro}

\begin{figure}[!t]
\centering
    \includegraphics[width=0.92\linewidth]{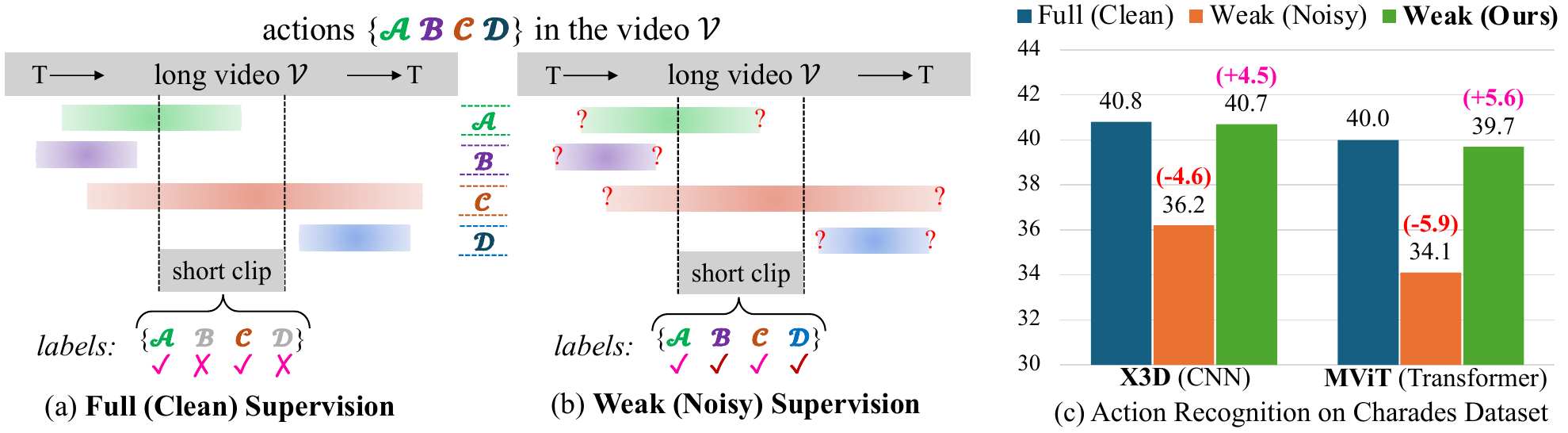}
    \caption{
    (a): End-to-end training action recognition models on long videos by sampling short clips, which requires full supervision of action categories and intervals; 
    (b): Similar to (a), but the action interval annotations are unavailable. Such weak supervision will introduce noisy action labels that hurt model training;
    (c): Models trained under noisy weak supervision exhibit significant performance degradation compared to full supervision training. With our AdaptFocus framework, models can alleviate the influence of noises, and achieve comparable results to the models trained under clean full supervision.
    }
\label{fig:motivation}
\vspace{-0.40cm}
\end{figure}

The long-video action recognition~\cite{wang2018non, feichtenhofer2019slowfast, fan2021multiscale} aims to recognize human actions in long videos, which is fundamental and challenging for understanding complex human activities in real life.
To ensure the high robustness and discriminability of the action recognition models, end-to-end training models on long-video data is necessary.
However, due to the high demand for memory and computation when processing long videos, existing works typically end-to-end train models by trimming short clips from long videos. 
As depicted in Figure~\ref{fig:motivation}(a), this approach requires precise annotations of the action’s start and end timestamps to accurately annotate which actions are contained within the trimmed clips, which are expensive to collect.
Therefore, in this work, we explore end-to-end learning for long-video action recognition in a weakly supervised manner, where action interval annotations are not accessible. 

Solving this weakly supervised long-video action recognition task under an end-to-end framework is non-trivial, as it faces a dilemma between the weak annotation and end-to-end training demand (i.e., the end-to-end training only allows clip-level processing while the weak annotation cannot provide precise supervision). 
Notably, previous works addressing weakly supervised long-video tasks (e.g., action localization~\cite{zhang2021cola, ju2022adaptive}) typically adopt a multi-instance learning (MIL) framework~\cite{ren2023proposal, luo2020weakly, zhang2020multi}, which treats the pre-extracted features of each long video as a bag. However, the MIL framework is not applicable for end-to-end training, since loading very long videos in memory is infeasible.
Alternatively, in Figure~\ref{fig:motivation}(b), we illustrate a straightforward approach for our setting, where all actions in the video-level category set are used as the supervised labels for the trimmed clips during end-to-end training. 
However, as shown in Figure~\ref{fig:motivation}(c), such weakly supervised training significantly degrades the recognition performance of both CNN-based and Transformer-based models. 
This degradation is caused by noisy action labels, i.e., some actions present in the long video do not appear in the trimmed clip.
To this end, this work aims to answer the question:
\begin{quote}
\vspace{-0.25cm}
Can action recognition models be effectively end-to-end trained on long-video datasets under weak supervision?
\vspace{-0.25cm}
\end{quote}

This problem can be treated as a noisy learning problem~\cite{natarajan2013learning, wu2021class2simi}. Inspired by self-paced learning~\cite{DBLP:conf/nips/KumarPK10, DBLP:conf/icml/BengioLCW09}, 
this work proposes a novel \textbf{AdaptFocus} framework, which guides models to \textbf{adapt}ively \textbf{focus} on action clips with cleaner supervision in long videos, during weakly supervised end-to-end training. 
For each action that appears in a long video, our AdaptFocus firstly estimates the temporal position where the action is most likely to occur, and the corresponding spike-actionness (i.e., the highest probability of action occurrence). 
Based on this, the AdaptFocus further selects action clips with cleaner supervision for end-to-end training, according to the following two intuitions.
First, for a new clip trimmed from the video, if its actionness of a certain action is close to the estimated spike-actionness of the action, the models can be more confident that this clip contains the action. 
Second, if clips at some positions exhibit spike-actionness, their neighboring clips are more likely to contain the same actions. 
In this way, as demonstrated in Figure~\ref{fig:motivation}(c), 
although video-level weak supervision introduces noise to the labels of trimmed clips,
our AdaptFocus framework effectively mitigates the performance drop caused by the noisy labels, achieving remarkable results that are comparable to the models trained under full supervision. 

In addition, as another contribution, the success of end-to-end training long-video action recognition models under weak supervision, motivates a new feature extraction pipeline for downstream weakly supervised long-video tasks (e.g., action segmentation~\cite{farha2019ms, bahrami2023much}). 
Specifically, these tasks usually require complex temporal modeling over the whole long videos, making end-to-end training impractical.
Current approaches typically build models upon frozen long-video features, which are pre-extracted by action recognition models pre-trained on short-video datasets.
This leads to a domain discrepancy~\cite{kahatapitiya2023weakly, zhao2023re2tal} between the pre-trains and target downstream long-video datasets.
With our proposed AdaptFocus framework, action recognition extractors can be end-to-end trained on long-video datasets in a weakly supervised manner.
This forms a new feature extraction pipeline that significantly boosts the performance of existing state-of-the-arts on three long-video tasks: temporal sentence grounding~\cite{zheng2022weakly, wang2022negative}, complex activity recognition~\cite{hussein2019timeception, yu2020rhyrnn}, and action segmentation~\cite{farha2019ms, bahrami2023much}.

The contributions of this work are three-fold:
\begin{itemize}[itemsep=0pt, topsep=0pt, parsep=0pt, partopsep=0pt]

\item We propose AdaptFocus, the first weakly supervised framework for efficiently end-to-end training action recognition models on long videos;

\item Experiments across three long-video datasets and six action recognition models validate the effectiveness of AdaptFocus. 
Remarkably, on the two largest datasets, AdaptFocus enables models trained under weak supervision to achieve comparable results to those trained under full supervision;

\item A new weakly supervised feature extraction pipeline is established for long-video datasets, which yields significant performance improvements in three downstream weakly supervised long-video tasks.

\end{itemize}

\section{Related Works}
\label{sec:related_works}

\subsection{Action Recognition in Long Videos}

Significant progress has been made in video action recognition~\cite{wang2018non, carreira2017quo, zhu2020comprehensive, wang2013action, karpathy2014large, girdhar2017actionvlad, wang2016temporal, zhou2018temporal, lin2019tsm, neimark2021video}, which primarily focuses on well-structured short videos~\cite{carreira2017quo, goyal2017something}, while real-world scenarios~\cite{wu2021towards, zhao2017temporal, li2022bridge} often demand the analysis of long videos~\cite{damen2020epic, tang2019coin}.
Current efforts in long-video action recognition~\cite{feichtenhofer2019slowfast, fan2021multiscale, hussein2019timeception, zhou2023twinformer} are primarily directed along two paths. The first emphasizes robust feature representation through end-to-end training on long videos~\cite{wang2018non, feichtenhofer2019slowfast, fan2021multiscale, wu2019long, feichtenhofer2020x3d, pang2021pgt}. However, this approach relies on costly action interval annotations, following the ``trimming then training'' paradigm.
The second path explores the modeling of long-term temporal relations~\cite{hussein2019timeception, zhou2023twinformer, yu2020rhyrnn, strafforello2023current}. Yet, it faces challenges due to substantial computation and memory requirements, leading to a reliance on pre-extracted features of long videos. Attempts have been made to bridge these two directions, such as LFB~\cite{wu2019long}, MeMViT~\cite{wu2022memvit} and PGT~\cite{pang2021pgt}, which employ feature banks or intricate training methods to enhance long-range context modeling with end-to-end training. However, the dependency on action interval annotations persists, and the temporal scope of long-term modeling remains limited.

Without using the expensive action interval annotations, this work proposes the first weakly supervised end-to-end framework for training action recognition models on long videos. This is very challenging because the huge cost of end-to-end training only allows to processing short clips of long videos, while the weak annotation cannot provide accurate supervision.

\subsection{Weakly Supervised Long-video Understanding}

The field of weakly supervised long-video action understanding encompasses various tasks, e.g., action localization~\cite{zhang2021cola, ju2022adaptive} and sentence grounding~\cite{zheng2022weakly, wang2022negative}. 
These fine-grain tasks usually require complex temporal modeling over the whole long videos. 
Thus pre-extracted long-video features are used due to the unaffordable cost of end-to-end training. Under the weakly supervised setting, many practices (e.g., multiple instance learning~\cite{ren2023proposal, luo2020weakly, zhang2020multi}) are developed for these non-end2end tasks, where features of the whole long videos are accessible without noisy supervision involved. 
However, in our different end-to-end learning setting, these practices are undesirable when developing a trimming-based fully-trained framework, where only short clips are allowed to be end-to-end processed and noisy supervision exists.

In addition, the extractors used in these weakly supervised long-video tasks are action recognition models trained on short videos, making the extracted features suffer domain discrepancy. 
One potential solution for this is using existing foundation models (e.g., VideoMAE~\cite{tong2022videomae}) trained on large-scale short videos to obtain long-video features, however, the scale of short-video pre-training data and its inconsistency with the domain of long-video data remain unresolved issues.
This motivates various works~\cite{kahatapitiya2023weakly, zhao2023re2tal, pang2021pgt, liu2020progressive, cheng2022tallformer, liu2022empirical, xu2021boundary, luo2022exploring, zhang2022unsupervised, alwassel2021tsp, kang2023soft} to mitigate the feature domain discrepancy by training or finetuning their models on long-video datasets. Despite their contributions, these methods often cater to specific tasks (e.g., BSP~\cite{xu2021boundary} and SoLa~\cite{kang2023soft}), use additional annotations (e.g., TSP~\cite{alwassel2021tsp}), or rely on intricate training strategies (e.g., PGT~\cite{pang2021pgt}). 
This work proposes the first end-to-end framework for training recognition models under weak supervision, which could serve as generic and stronger backbones for many weakly supervised long-video tasks, without using extra annotations.

\subsection{Self-paced Learning}

Self-paced learning (SPL)~\cite{DBLP:conf/nips/KumarPK10} and curriculum learning~\cite{DBLP:conf/icml/BengioLCW09} emphasize learning from simple to complex tasks.
They suggest a structured approach to present training data, starting with clean (easy) samples and progressively introducing noisier (hard) data to enhance model generalization. Inspired by SPL, subsequent research~\cite{DBLP:conf/nips/JiangMYLSH14, DBLP:conf/mir/LiangJMH17, chen2020self, ma2017self, jiang2018mentornet, DBLP:conf/aaai/ZhangWCZL20} has explored various strategies to quantify the complexity and noise within data. 

This work focuses on weakly supervised end-to-end training on long videos, where the supervision involves noise. Inspired by the insight of selecting cleaner supervision in SPL, our proposed AdaptFocus, integrates an adaptive focus mechanism to select actions and clips within long videos that are least contaminated by label noise for end-to-end training models.

\section{AdaptFocus Learning Framework}
\label{sec:method}

\subsection{Problem formulation}
Assuming we have a long video, denoted as $V=\{s_1, s_2, \ldots, s_t, \ldots, s_T\}$, where $s_t$ represents the $t$-th clip containing multiple frames. The corresponding video-level action category labels are $Y=[y^0, \ldots, y^k, \ldots, y^K]\in \{0,1\}^K$, where $K$ represents the number of action classes in the dataset, and $y^k=1$ indicates that the $k$-th action exists in the long video $V$. 
During training, a single short clip (e.g., the $t$-th clip $s_t$) will be trimmed from the long video for end-to-end training action recognition models. Without the full supervision of action interval (start and end timestamps) annotations, models cannot determine which actions are present in the sampled clip. 

A straightforward approach to address this challenge is to treat all actions in the video as labels for the sampled clip, as shown in Figure~\ref{fig:motivation}(b). Under such weak supervision, the supervised training loss on the $t$-th clip of long video $V$ is formulated as:
\begin{equation}
\setlength\abovedisplayskip{3pt}
\setlength\belowdisplayskip{-1pt}
\begin{split}
\mathcal{L}^{noisy} = \mathcal{L}_{in\_vid} + \mathcal{L}_{out\_vid}
= \sum_{k=1}^{K}-y^k\log p_t^k + \sum_{k=1}^{K}-(1-y^k)\log (1-p_t^k),  
\end{split}
\end{equation}

where $p_t^k$ is the sigmoid-activated prediction of the $k$-th action. $\mathcal{L}_{in\_vid}$ and $\mathcal{L}_{out\_vid}$ are the loss terms for positive and negative labels, respectively. 
However, as some actions in the video do not occur in the sampled clip, these actions will serve as noise in the first loss term $\mathcal{L}_{in\_vid}$, leading to a significant degradation in the performance of the trained models (as shown in Figure~\ref{fig:motivation}(c)).

To this end, the goal of this work is to mitigate the effect of noisy actions within the labels of short clips under weak supervision. For clarity, we will not show the second loss term $\mathcal{L}_{out\_vid}$ in the following, where no noisy action is involved (since the video-level negatives are not present in any of the clips).


\subsection{Overview}

\begin{wrapfigure}{r}{0.47\textwidth}
\vspace{-0.2cm}
  \centering
   \includegraphics[width=1.0\linewidth]{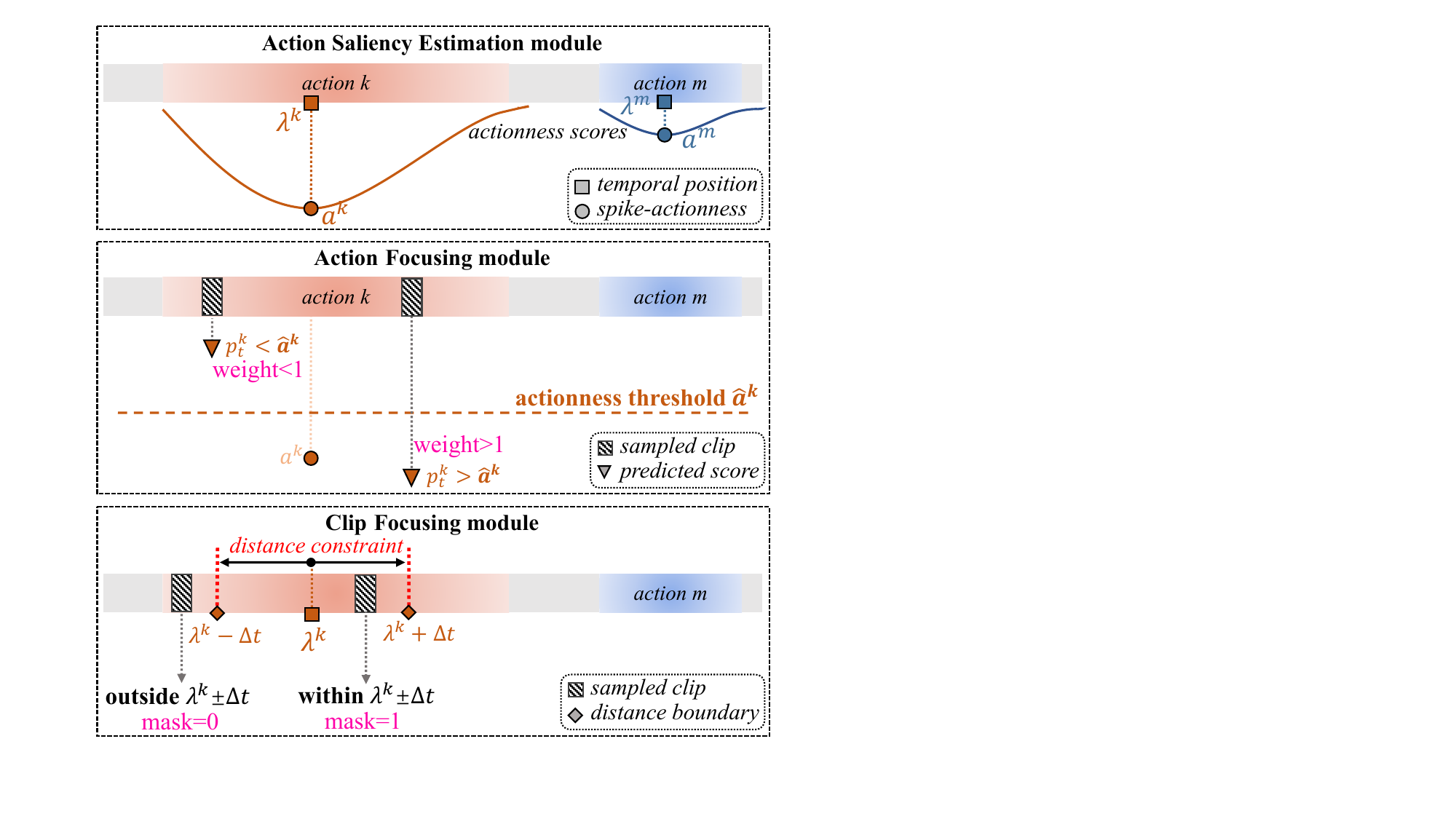}
   \caption{Illustrations of the three modules in the proposed AdaptFocus framework.}
\label{fig:method}
\vspace{-0.6cm}
\end{wrapfigure}
This work proposes the AdaptFocus framework to end-to-end train action recognition models on long videos, in a weakly supervised manner. The challenge is that during end-to-end training, only a short clip can be sampled, while the video-level weak annotations cannot provide precise supervision. This problem can be treated as a noisy learning problem.
Inspired by self-paced learning~\cite{DBLP:conf/nips/KumarPK10}, 
where model optimization begins with the simplest/cleanest data and gradually involves more difficult/noisy data, 
our AdaptFocus framework adaptively focuses on actions and clips that contain cleaner supervision.

As illustrated in Figure \ref{fig:method}, the proposed AdaptFocus framework consists of three modules, i.e., Action Saliency Estimation (ASE) module, Action Focusing module, and Clip Focusing module.
With only video-level action labels for weak supervision, the challenge lies in determining which actions are present in the trimmed clips for supervised end-to-end training. 
To solve this, for each action that appears in the long video, the ASE module is designed to estimate the temporal position where the action is most likely to occur, along with the action's spike-actionness at that position.
According to the estimated spike-actionness of actions, for a newly sampled clip in the long video, the Action Focusing module determines whether these actions exist in the clip. 
Besides, since the actions have continuity along the temporal dimension, the Clip Focusing module will focus on clips whose temporal positions are close to the temporal positions with estimated spike-actionness.
The integration of the Action Focusing and Clip Focusing modules effectively minimizes the impact of noisy actions, leading to a more reliable model. This enhanced model confidence, in turn, improves the reliability of the ASE module's estimations in subsequent iterations.

\vspace{-0.2cm}
\subsection{Action Saliency Estimation (ASE) module}
\vspace{-0.1cm}
For each action (e.g., the $k$-th action) that occurs in the long video, the ASE module estimates the most salient temporal position $\lambda^k$ of the action, and the corresponding spike-actionness value $a^k$ at the position $\lambda^k$.

\noindent\textbf{- Naive Estimation.} To achieve this, we apply an action classifier to all clips of the video. For the $k$-th action in the video, the temporal position and classification score of the clip that has the highest score among all clips, are estimated as the position $\lambda^k$ and spike-actionness $a^k$. This process can be formulated as:
\setlength{\abovedisplayskip}{2pt}
\setlength{\belowdisplayskip}{2pt}
\begin{align}
\lambda^k = \mathop{\arg\max}_{t\in [1,T]} p_t^k, \quad \quad a^k = \max_{t\in [1,T]} p_t^k,
\end{align}
where $T$ is the number of clips in the video, and $p_t^k$ is classification score of the $k$-th action in the $t$-clip.

\noindent\textbf{- Online Estimation.} The above estimation requires inference on all clips of the video for every training iteration, which is inefficient. 
During the trimming-based end-to-end training, a single short clip is randomly trimmed from each long video sample.
As the training progresses, the number of sampled clips will eventually be sufficient for estimation.
Thus, the following online estimation is used to replace the above estimation process:
\setlength{\abovedisplayskip}{2pt}
\setlength{\belowdisplayskip}{2pt}
\begin{align}
\label{online_estimation}
\lambda^k = t \ \ \text{if } \ p_t^k>a^k, \quad \quad a^k = \max(a^k,p_t^k),
\end{align}
which means that the position $\lambda^k$ and spike-actionness $a^k$ will be updated, only when the currently sampled clip shows more saliency of the action than the previous most-salient clip (i.e., $p_t^k>a^k$). 
Using the online estimation, our ASE module only needs to save $2\times D$ values (i.e., $D$ is the number of action instances in the dataset) in memory, and this cost is negligible.

\subsection{Action Focusing module}
\vspace{-0.1cm}
For each action that appears in the long video, our Action Focusing module is proposed to determine whether it appears in the sampled clip. We use the spike-actionness of the action estimated by the ASE module as a reference. 
If the classification score $p_t^k$ of the $t$-th clip is close to the spike-actionness $a^k$ of the $k$-th action, the model can be more confident that this action appears in the clip, and vice versa. 
To this end, our action-focus loss $\mathcal{L}_{in\_vid}^{action}$ is formulated to adjust the loss weight of each action on the sampled clip: 
\setlength{\abovedisplayskip}{0pt}
\setlength{\belowdisplayskip}{0pt}
\begin{gather}
\mathcal{L}_{in\_vid}^{action} = - \sum_{k=1, y^k=1}^{K} \mathcal{W}(p_t^k, \hat{a}^k) \cdot \log p_t^k, 
\label{eq:action_loss}
\end{gather}
where $\hat{a}^k=\theta \cdot a^k, \ \theta\in(0,1)$. $\hat{a}^k$ is the actionness threshold based on the spike-actionness $a^k$. 
And $\mathcal{W}$ is a soft weighting function, which assigns real-valued weight that reflects the importance of each action class during training.
Inspired by self-paced learning~\cite{DBLP:conf/nips/KumarPK10}, our weighting function $\mathcal{W}$ is defined as follows:
\setlength{\abovedisplayskip}{0pt}
\setlength{\belowdisplayskip}{0pt}
\begin{gather}
\large
\mathcal{W}(p_t^k,\hat{a}^k) = 
\begin{cases}\alpha\cdot e^{(p_t^k-\hat{a}^k)}, & p_t^k>=\hat{a}^k, \\
e^{-\beta \cdot {(\hat{a}^k-p_t^k)}}, & p_t^k<\hat{a}^k, \\
\end{cases}
\label{eq:weight_func}
\end{gather}
where factors $\alpha>=1$ and $\beta>0$ are used to control the scaling of weights (see the visualization in Figure~\ref{fig:ablation_weight_func_sm} of the \textbf{Appendix}). 
When the action score $p_t^k$ of the current clip is larger than the actionness threshold $\hat{a}^k$, the value of the weighting function $\mathcal{W}$ will be larger than $1.0$, and vice versa. 
We analyze the effect of using different weighting functions in Section~\ref{ablation_weighting_func} of the \textbf{Appendix}.

\subsection{Clip Focusing module}
In addition, we propose a Clip Focusing module to improve the attention on clips in which actions are more likely to exist. 
In this way, the model will be exposed to less noisy data during training. 
The model can in turn help identify which clips need to be focused on in the next training iterations. 

Given the temporal position $\lambda^k$ where the $k$-th action is the most salient, the model is constrained to focus on clips whose positions are close to the temporal position $\lambda^k$. 
Since those clips are more likely to have the same action due to the temporal continuity of actions. Such constraint is formulated as the following clip-focus loss $\mathcal{L}_{in\_vid}^{clip}$:
\setlength{\abovedisplayskip}{0pt}
\setlength{\belowdisplayskip}{0pt}
\begin{gather}
\mathcal{L}_{in\_vid}^{clip} = - \sum_{k=1, y^k=1}^{K} \mathcal{M}(t, T, \lambda^k, \gamma) \cdot \log p_t^k,
\label{eq:clip_loss}
\end{gather}
where $t$ is the position of the current clip, and $T$ is the length of the video. 
Similar to the binary weighting function~\cite{DBLP:conf/nips/KumarPK10} in self-paced learning, $\mathcal{M}$ is a masking function.
When the normalized temporal distance between the current clip and the clip with spike-actionness is smaller than the factor $\gamma$, the clip is selected for training, and the value of the masking function $\mathcal{M}$ is $1$, otherwise it is $0$. The masking function $\mathcal{M}$ is defined as follows:
\setlength{\abovedisplayskip}{2pt}
\setlength{\belowdisplayskip}{2pt}
\begin{gather}
\mathcal{M}(t,T,\lambda^k, \gamma) =  
\begin{cases}
1, & 2\cdot|t-\lambda^k|/T<=\gamma, \\
0, & 2\cdot|t-\lambda^k|/T>\gamma, \\
\end{cases}
\end{gather}
where the factor $\gamma \in [0,1]$ controls the scale of distance constraint, whose value is the ratio of the current epoch to the total training epoch, gradually increasing from $0$ to $1$. In this way, more clips in the video will be gradually covered during training.

To effectively guide models to focus on cleaner data, the weighting function $\mathcal{W}$ in action-focus loss $\mathcal{L}_{in\_vid}^{action}$ (Eq.~\ref{eq:action_loss}) and the masking function $\mathcal{M}$ in clip-focus loss $\mathcal{L}_{in\_vid}^{clip}$ (Eq.~\ref{eq:clip_loss}) can be simultaneously applied:
\setlength{\abovedisplayskip}{-2pt}
\setlength{\belowdisplayskip}{0pt}
\begin{gather}
\mathcal{L}_{in\_vid}^{AdaptFocus} = -\sum_{k=1, y^k=1}^{K} \mathcal{M}(t,T,\lambda^k, \gamma) \cdot \mathcal{W}(p_t^k, \hat{a}^k)\cdot\log p_t^k.
\end{gather}
Thus, the final loss for weakly supervised end-to-end training on long videos is formulated as follows:
\setlength{\abovedisplayskip}{2pt}
\begin{gather}
\mathcal{L}^{Ours} = \mathcal{L}_{in\_vid}^{AdaptFocus} + \mathcal{L}_{out\_vid}.
\end{gather}

\section{Experiments}
\label{sec:exps}
In this section, we first present the implementation details of the AdaptFocus framework. 
Then we show the effectiveness of AdaptFocus (which is a weakly supervised training method) by comparing it with noisy training and fully supervised (oracle) training. 
Later, ablation studies on AdaptFocus are elaborated. 
Finally, we propose a new weakly supervised feature extraction pipeline using AdaptFocus, 
which significantly boosts state-of-the-arts on three weakly supervised long-video tasks.

\subsection{Implementation Details}
The factors $\theta$, $\alpha$, and $\beta$ are empirically set to $0.75$, $5.0$, and $3.0$, respectively. All these factors are the same across different datasets and models. Ablation studies on these three factors are provided in Section~\ref{ablation_parameter} of \textbf{Appendix}. The initial values of position $\lambda$ and spike-actionness $a$ for each action instance are set to $0$. 
The model is initially trained under weak supervision without AdaptFocus for $20\%$ of the total iterations, where the positions $\lambda$ and spike-actionness $a$ of actions are updated by Online Estimation (see Eq.~\ref{online_estimation}), making them more reliable when starting the learning of the AdaptFocus framework. 
Other hyper-parameters (e.g., learning rate, batch size) vary with different models and datasets, and we follow the same settings as their public codebases for fair comparison. During training, $F$ frames with a stride $\tau$ (denoted as $F\times \tau$) are sampled for each clip. For testing, following previous practice, $\textit{view}_t\times \textit{view}_s$ clips are evenly sampled from a long video, where $\textit{view}_t$ is the number of temporal clips and $\textit{view}_s$ is the number of spatial crops. The classification scores of these clips are averaged into the final video-level classification score. All experiments are conducted on $4$ NVIDIA A6000 GPUs for about 12 hours.

\subsection{Effectiveness of AdaptFocus Framework}
Three popular long-video datasets, i.e., Charades~\cite{sigurdsson2016hollywood}, Breakfast~\cite{kuehne2014language}, and MultiThumos~\cite{yeung2018every}, are used. To comprehensively validate the effectiveness of our AdaptFocus framework, both CNN-based methods (e.g., SlowFast~\cite{feichtenhofer2019slowfast}) and Transformer-based methods (e.g., MViT~\cite{fan2021multiscale}) are evaluated. 

\noindent\textbf{- Charades dataset}~\cite{sigurdsson2016hollywood} contains $157$ daily actions, with $7811$ videos for training and $1814$ videos for testing. On this dataset, $10\times 3$ views are used for testing. Table~\ref{tab:results_on_charades} presents the results of nine video-based models (Inception-I3D~\cite{carreira2017quo}, ResNet50-I3D~\cite{wang2018videos}, Nonlocal-I3D~\cite{wang2018non}, SlowFast-R50~\cite{feichtenhofer2019slowfast}, X3D-L~\cite{feichtenhofer2020x3d}, and four MViT variants~\cite{fan2021multiscale}), with different architectures and sampling strategies. All these models are pretrained on Kinetics-400~\cite{carreira2017quo}, except for MViT-B-24, which is pretrained on Kinetics-600~\cite{carreira2018short}. 

In Table~\ref{tab:results_on_charades}, the second column, i.e., \myGray{Full (Clean)}, shows the results of fully supervised end-to-end training, where action interval annotations are used to obtain the clean labels for trimmed short-clips. However, such annotations are expensive to collect. 
The third column, i.e., \myRed{Weak (Noisy)}, of Table~\ref{tab:results_on_charades} presents the results of weakly supervised end-to-end training, where the video-level action categories are treated as the labels for trimmed-clips. 
Under this setting, some noisy action labels will be introduced into the trimmed clips, leading to a significant decline in recognition performance compared to full supervision. 
To solve this, our proposed AdaptFocus framework adaptively focuses on actions and clips that contain cleaner supervision. 
The results in the last column, i.e., \myPink{Weak (Ours)}, of Table~\ref{tab:results_on_charades} show that our AdaptFocus enables weakly supervised models to achieve comparable results to the fully supervised models (i.e., oracle models), demonstrating its superiority in avoiding the interference of noisy actions under weak supervision.

\begin{figure}
\centering
\begin{minipage}{0.49\linewidth}
\centering
\setlength{\abovecaptionskip}{0cm}
\setlength{\belowcaptionskip}{0.0cm}
\captionof{table}{\small{Action recognition results on \textbf{Charades} dataset. The mean average precision (mAP) metric is used. $F\times \tau$ denotes \textit{the number of frames} $\times$ \textit{stride between frames} of clips. \myGray{Full (Clean)} means that the action interval annotations are used. \myRed{Weak (Noisy)} denotes that the all video-level actions are used as the labels of trimmed clips. \myPink{Weak (Ours)} denotes that the proposed AdaptFocus is applied during end-to-end training under weak supervision.}}
\resizebox{1.0\textwidth}{!}
{
\begin{tabular}{l|c|c|c}
\specialrule{0.9pt}{0pt}{0pt}
\multicolumn{1}{c|}{\multirow{2}{*}{Model, $F\times \tau$}}  & \multicolumn{3}{c}{mAP (\%) (\textit{views:} $10\times 3$)} \\
\cline{2-4}
\multicolumn{1}{c|}{}       & \myGray{\small{Full (Clean)}}    & \myRed{\small{Weak (Noisy)}}    & \myPink{\small{Weak (Ours)}} \\
\specialrule{0.7pt}{0pt}{0pt}
Inception-I3D, $32\times3$  &  \myGray{32.9}  &  28.7 \textcolor{red}{(-4.2)}  &  \textbf{32.8} \textbf{\myPink{(+4.1)}} \\
\hline
ResNet50-I3D,  $32\times3$  &  \myGray{31.8}  &  30.2 \myRed{(-1.6)}  &  \textbf{31.5} \textbf{\myPink{(+1.3)}} \\
\hline
Nonlocal-I3D,  $32\times3$  &  \myGray{33.5}  &  30.9 \myRed{(-2.6)}  &  \textbf{33.5} \textbf{\myPink{(+2.6)}} \\
\hline
SlowFast-R50,  $16\times8$  &  \myGray{38.9}  &  34.3 \myRed{(-4.6)}  &  \textbf{38.6} \textbf{\myPink{(+4.3)}} \\
\hline
X3D-L,         $32\times3$  &  \myGray{40.8}  &  36.2 \myRed{(-4.6)}  &  \textbf{40.7} \textbf{\myPink{(+4.5)}} \\
\hline
MViT-B,        $16\times4$  &  \myGray{40.0}  &  34.1 \myRed{(-5.9)}  & \textbf{39.7} \textbf{\myPink{(+5.6)}}  \\
MViT-B,        $32\times3$  &  \myGray{44.3}  &  38.3 \myRed{(-6.0)}  & \textbf{44.1} \textbf{\myPink{(+5.8)}}  \\
MViT-B,        $64\times3$  &  \myGray{46.3}  &  42.2 \myRed{(-4.1)}  & \textbf{45.9} \textbf{\myPink{(+3.7)}}  \\
MViT-B-24,     $32\times3$  &  \myGray{46.9}  &  42.0 \myRed{(-4.9)}  & \textbf{46.8} \textbf{\myPink{(+4.8)}}  \\
\specialrule{0.9pt}{0pt}{0pt}
\end{tabular}
}
\label{tab:results_on_charades}
\end{minipage}
\hfill
\begin{minipage}{0.49\linewidth}
\begin{minipage}{1.0\linewidth}
\centering
\setlength{\abovecaptionskip}{0.cm}
\setlength{\belowcaptionskip}{-0.2cm}
\captionof{table}{\small{Action recognition results on \textbf{Breakfast} and \textbf{MultiThumos} datasets.}}
\resizebox{1.0\textwidth}{!}
{
\begin{tabular}{l|c|c|c}
\specialrule{0.9pt}{0pt}{0pt}
\multicolumn{1}{c|}{\multirow{2}{*}{Model, $F\times \tau$}} & \multicolumn{3}{c}{mAP (\%) (\textit{views:} $30\times 1$)} \\
\cline{2-4}
\multicolumn{1}{c|}{}       & \myGray{\small{Full (Clean)}}    & \myRed{\small{Weak (Noisy)}}    & \myPink{\small{Weak (Ours)}} \\
\specialrule{0.9pt}{0pt}{0pt}
\addlinespace[0.5ex]

\specialrule{0.5pt}{0pt}{0pt}
\multicolumn{4}{c}{Breakfast dataset} \\
\hline
Inception-I3D, $32\times3$  &  \myGray{59.6}  &  55.5 \textcolor{red}{(-4.1)}  &  \textbf{59.3} \textbf{\myPink{(+3.8)}} \\
\hline
MViT-B-24,     $32\times3$  &  \myGray{68.3}  &  65.2 \myRed{(-3.1)}  & \textbf{68.1} \textbf{\myPink{(+2.9)}}  \\
\specialrule{0.5pt}{0pt}{0pt}
\addlinespace[0.5ex]

\specialrule{0.5pt}{0pt}{0pt}
\multicolumn{4}{c}{MultiThumos dataset} \\
\hline
Inception-I3D, $32\times3$  &  \myGray{78.9}  &  74.9 \textcolor{red}{(-4.0)}  &  \textbf{75.4} \textbf{\myPink{(+0.5)}} \\
\hline
MViT-B-24,     $32\times3$  &  \myGray{83.2}  &  78.9 \myRed{(-4.3)}  & \textbf{81.6} \textbf{\myPink{(+2.7)}}  \\
\specialrule{0.5pt}{0pt}{0pt}
\end{tabular}
}
\label{tab:results_on_breakfast_and_multithumos}
\end{minipage}
\vspace{0.2cm}

\vfill
\begin{minipage}{1.0\linewidth}
\centering
\setlength{\abovecaptionskip}{0cm}
\setlength{\belowcaptionskip}{0cm}
\caption{\small{Ablation on Clip Focusing and Action Focusing modules on Charades dataset.}}
\resizebox{0.75\textwidth}{!}
{
\begin{tabular}{c|c}
\specialrule{0.9pt}{0pt}{0pt}
Model & mAP (\%) \\
\hline
Weak (Noisy) & 34.1 \\
\textit{w/} Clip Focusing & 37.4 (+3.3)\\
\textit{w/} Action Focusing & 38.3 (+4.2)\\
AdaptFocus & \textbf{39.7 (+5.6)} \\
\specialrule{0.9pt}{0pt}{0pt}
\end{tabular}
}
\label{tab:ab_on_clip_action_modules}
\end{minipage}
\end{minipage}
\vspace{-0.6cm}
\end{figure}

\noindent\textbf{- Breakfast Dataset}~\cite{kuehne2014language} consists of $48$ cooking actions across a total of $1712$ videos. Following the split used in~\cite{hussein2019timeception}, $1357$ videos are used for training and $355$ for testing. 
Experimental results, as presented in Table~\ref{tab:results_on_breakfast_and_multithumos}, also demonstrate that weakly supervised models employing our AdaptFocus framework achieve results that are comparable to their fully supervised counterparts. This improvement is evident in both CNN-based and Transformer-based models.

\noindent\textbf{- MultiThumos Dataset}~\cite{yeung2018every} extends the Thumos dataset~\cite{idrees2017thumos} by densely annotating $413$ long videos with $65$ sport action classes. It contains $200$ videos for training and $213$ for testing. 
The fine-grained nature of the actions in this dataset, many of which last less than $2$ seconds, presents a challenge for models trained under weak supervision. 
However, as indicated in Table~\ref{tab:results_on_breakfast_and_multithumos}, the models integrated with AdaptFocus show notable improvement over those directly trained in weak and noisy conditions (i.e., \myRed{Weak (Noisy)}), underlining the efficacy of our proposed framework. 

\subsection{Ablation Studies}

In this section, we present ablation studies on the designs of the AdaptFocus framework. These experiments are conducted on the Charades dataset using the MViT-B model~\cite{fan2021multiscale} with a $16\times 4$ frame sampling protocol.

\noindent\textbf{- Clip/Action Focusing modules.} The AdaptFocus framework's resistance to noise primarily stems from the Clip Focusing and Action Focusing modules. Table~\ref{tab:ab_on_clip_action_modules} shows the ablation results of these two modules. Compared with the baseline, i.e., \textit{Weak (Noisy)}, $3.3\%$ improvement in mAP is obtained when using only the Clip Focusing module. Similarly, there is a $4.2\%$ improvement in mAP when using only the Action Focusing module. Notably, by using both Clip Focusing and Action Focusing modules, i.e., \textit{AdaptFocus}, the improvement can be further boosted, which indicates a synergistic effect between the two modules.

\begin{figure}
\centering
\begin{minipage}{0.49\linewidth}
\centering
\captionof{table}{\small{Temporal Sentence Grounding on Charades.}}
\resizebox{1.0\linewidth}{!}
{
\begin{tabular}{c|c|c|c|c|c}
\specialrule{0.9pt}{0pt}{0pt}
\multicolumn{1}{c|}{\multirow{2}{*}{\makecell[c]{Super-\\vision}}} & \multicolumn{1}{c|}{\multirow{2}{*}{Model}} & \multicolumn{1}{c|}{Extr-} & \multicolumn{3}{c}{R1@IoU=0.5 (\%)} \\
\cline{4-6}
\multicolumn{1}{c|}{} & \multicolumn{1}{c|}{} & \multicolumn{1}{c|}{actor} & \myGray{Full-FEAT} & \myBlue{Kinc-FEAT}   & \myPink{Weak-FEAT} \\
\specialrule{0.9pt}{0pt}{0pt}

\addlinespace[0.5ex]
\specialrule{0.5pt}{0pt}{0pt}
\multicolumn{1}{c|}{\multirow{2}{*}{Weak}} & \multicolumn{1}{c|}{\multirow{2}{*}{\makecell[c]{CPL}}} & I3D & \myGray{48.6} & 39.6 & \textbf{48.1 \myPink{(+8.5)}} \\
\cline{3-6}
\multicolumn{1}{c|}{} & \multicolumn{1}{c|}{} & MViT & \myGray{53.7} & 47.8 & \textbf{51.7 \myPink{(+3.9)}} \\
\specialrule{0.5pt}{0pt}{0pt}

\addlinespace[0.5ex]
\specialrule{0.5pt}{0pt}{0pt}
\multicolumn{1}{c|}{Semi-} & \multicolumn{1}{c|}{\multirow{2}{*}{\makecell[c]{D3G}}} & I3D & \myGray{45.5} & 41.7 & \textbf{45.3 \myPink{(+3.6)}} \\
\cline{3-6}
\multicolumn{1}{c|}{weak} & \multicolumn{1}{c|}{} & MViT & \myGray{49.3} & 46.0 & \textbf{49.1 \myPink{(+3.1)}} \\
\specialrule{0.5pt}{0pt}{0pt}

\addlinespace[0.5ex]
\specialrule{0.5pt}{0pt}{0pt}
\multicolumn{1}{c|}{\multirow{2}{*}{Full}} & \multicolumn{1}{c|}{\multirow{2}{*}{\makecell[c]{MMN}}} & I3D & \myGray{55.8} & 49.4 & \textbf{55.7 \myPink{(+6.3)}} \\
\cline{3-6}
\multicolumn{1}{c|}{} & \multicolumn{1}{c|}{} & MViT& \myGray{62.1} & 55.2 & \textbf{61.4 \myPink{(+6.2)}} \\
\specialrule{0.5pt}{0pt}{0pt}
\end{tabular}
}
\label{tab:tsg_partial}
\end{minipage}
\hfill
\begin{minipage}{0.49\linewidth}
\centering
\captionof{table}{\small{Complex Activity Recognition on Breakfast.}}
\resizebox{1.0\linewidth}{!}
{
\begin{tabular}{c|c|c|c|c}
\specialrule{0.9pt}{0pt}{0pt}
\multicolumn{1}{c|}{\multirow{2}{*}{Model}} & \multicolumn{1}{c|}{Extr-} & \multicolumn{3}{c}{mAP (\%)} \\
\cline{3-5}
\multicolumn{1}{c|}{} & \multicolumn{1}{c|}{actor} & \myGray{Full-FEAT} & \myBlue{Kinc-FEAT}   & \myPink{Weak-FEAT} \\
\specialrule{0.9pt}{0pt}{0pt}

\addlinespace[0.5ex]
\specialrule{0.5pt}{0pt}{0pt}
\multicolumn{1}{c|}{\multirow{2}{*}{MLP}} & I3D & \myGray{71.1} & 46.1 & \textbf{64.1 \myPink{(+18.0)}} \\
\cline{2-5}
\multicolumn{1}{c|}{} & MViT & \myGray{74.7} & 58.0 & \textbf{71.6 \myPink{(+13.6)}} \\
\specialrule{0.5pt}{0pt}{0pt}

\addlinespace[0.5ex]
\specialrule{0.5pt}{0pt}{0pt}
\multicolumn{1}{c|}{\multirow{2}{*}{\makecell[c]{Timeception}}} & I3D & \myGray{74.2} & 59.6 & \textbf{70.4 \myPink{(+10.8)}} \\
\cline{2-5}
\multicolumn{1}{c|}{} & MViT & \myGray{82.5} & 66.7 & \textbf{79.2 \myPink{(+12.5)}} \\
\specialrule{0.5pt}{0pt}{0pt}

\addlinespace[0.5ex]
\specialrule{0.5pt}{0pt}{0pt}
\multicolumn{1}{c|}{\multirow{2}{*}{\makecell[c]{GHRM}}} & I3D & \myGray{74.7} & 62.6 & \textbf{69.6 \myPink{(+7.0)}} \\
\cline{2-5}
\multicolumn{1}{c|}{} & MViT & \myGray{83.2} & 68.7 & \textbf{79.5 \myPink{(+10.8)}} \\
\specialrule{0.5pt}{0pt}{0pt}
\end{tabular}
}
\label{tab:lar_breakfast}
\end{minipage}
\vfill
\begin{minipage}{0.50\linewidth}
\setlength{\abovecaptionskip}{0cm}
\setlength{\belowcaptionskip}{0cm}
\captionof{table}{\small{Action Segmentation on Breakfast.}}
\small
\begin{center}
\resizebox{1.0\textwidth}{!}
{
\begin{tabular}{c|c|c|c|c|c}
\specialrule{0.9pt}{0pt}{0pt}
\multicolumn{1}{c|}{\multirow{2}{*}{\makecell[c]{Super-\\vision}}} & \multicolumn{1}{c|}{\multirow{2}{*}{Model}} & \multicolumn{1}{c|}{\multirow{2}{*}{Metric}} & \multicolumn{3}{c}{N runs (\%)} \\
\cline{4-6}
\multicolumn{1}{c|}{} & \multicolumn{1}{c|}{} & \multicolumn{1}{c|}{} & \myGray{Full-FEAT} & \myBlue{Kinc-FEAT}   & \myPink{Weak-FEAT} \\
\specialrule{0.9pt}{0pt}{0pt}

\addlinespace[0.5ex]
\specialrule{0.5pt}{0pt}{0pt}
\multicolumn{1}{c|}{\multirow{4}{*}{Weak}} & \multicolumn{1}{c|}{\multirow{4}{*}{\makecell[c]{POC}}} & MoF (avg) & \myGray{48.9} & 40.1 & \textbf{47.8 \myPink{(+7.7)}} \\
\cline{3-6}
\multicolumn{1}{c|}{} & \multicolumn{1}{c|}{} & MoF (max) & \myGray{49.6} & 42.4 & \textbf{49.6 \myPink{(+7.2)}}  \\
\cline{3-6}
\multicolumn{1}{c|}{} & \multicolumn{1}{c|}{} & IoU (avg) & \myGray{35.2} & 32.5 & \textbf{34.9 \myPink{(+2.4)}} \\
\cline{3-6}
\multicolumn{1}{c|}{} & \multicolumn{1}{c|}{} & IoU (max) & \myGray{37.6} & 33.5 & \textbf{37.6 \myPink{(+4.1)}} \\

\specialrule{0.5pt}{0pt}{0pt}

\addlinespace[0.5ex]
\specialrule{0.5pt}{0pt}{0pt}
\multicolumn{1}{c|}{\multirow{5}{*}{Full}} & \multicolumn{1}{c|}{\multirow{5}{*}{\makecell[c]{LT-\\Con-\\text}}} & F1@10 & \myGray{85.3} & 77.6 & \textbf{82.1 \myPink{(+4.5)}}  \\
\cline{3-6}
\multicolumn{1}{c|}{} & \multicolumn{1}{c|}{} & F1@25 & \myGray{82.1} & 72.6 & \textbf{79.0 \myPink{(+6.4)}} \\
\cline{3-6}
\multicolumn{1}{c|}{} & \multicolumn{1}{c|}{} & F1@50 & \myGray{71.9} & 60.1 & \textbf{67.5 \myPink{(+7.4)}} \\
\cline{3-6}
\multicolumn{1}{c|}{} & \multicolumn{1}{c|}{} & Edit & \myGray{81.7} & 77.0 & \textbf{78.3 \myPink{(+1.3)}} \\
\cline{3-6}
\multicolumn{1}{c|}{} & \multicolumn{1}{c|}{} & Acc & \myGray{82.5} & 74.2 & \textbf{78.0 \myPink{(+3.8)}} \\
\specialrule{0.5pt}{0pt}{0pt}
\end{tabular}
}
\end{center}
\label{tab:tas_breakfast}
\vspace{-0.3cm}
\end{minipage}
\hfill
\begin{minipage}{0.49\linewidth}
\centering
\small
\includegraphics[width=0.7\textwidth]{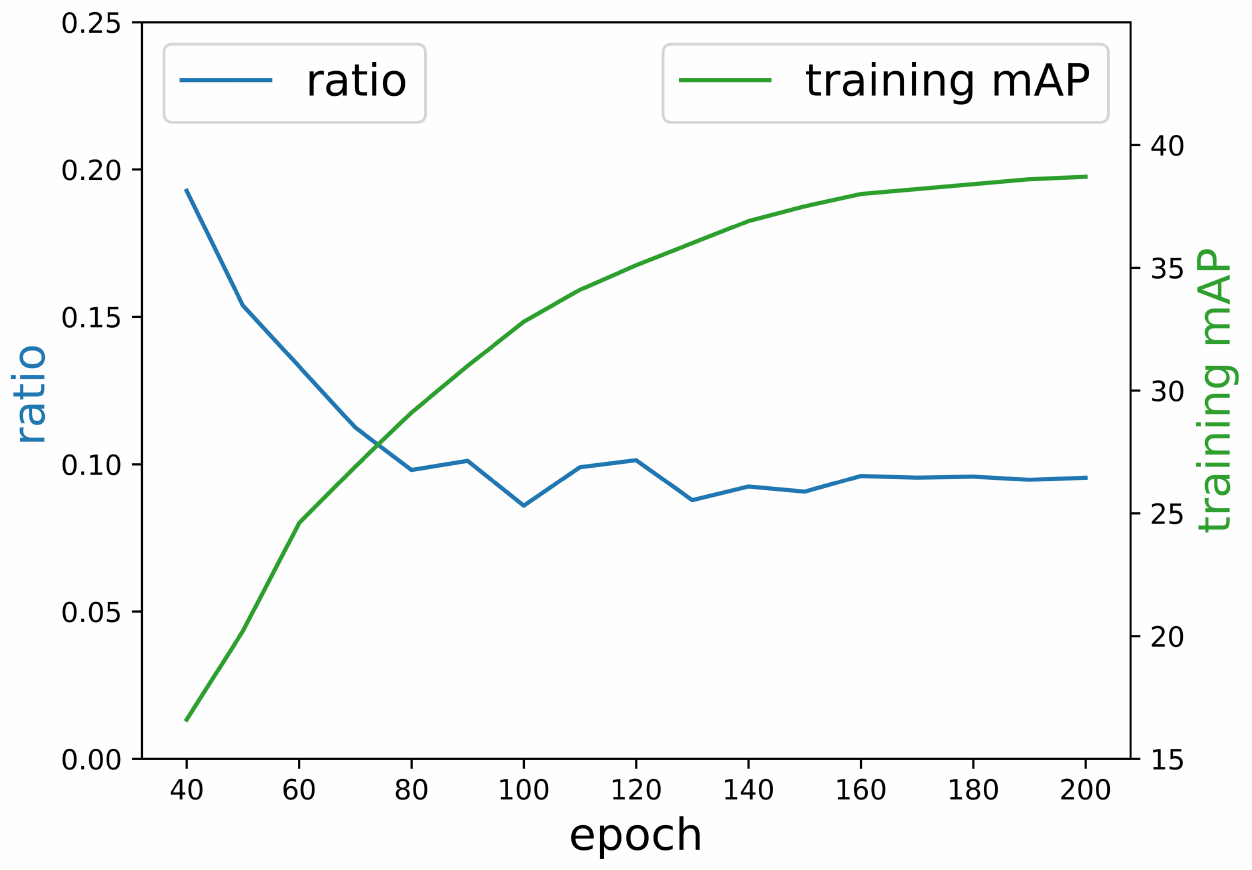}
\caption{\small{The changes of the ratio of scores exceeding the threshold in all predictions and the training mAP during optimization.}}
\label{fig:ablation_positive_weight_rate}
\end{minipage}%
\vspace{-0.60cm}
\end{figure}

\noindent\textbf{- Spike-actionness Estimation.} The effectiveness of the Clip/Action Focusing modules relies on the estimation of spike-actionness in the Action Saliency Estimation module. In our design, the estimated spike-actionness $a$ is continuously increasing during training, thus the actionness threshold $\hat{a}$ will increase accordingly. 
The Action Focusing module gives higher weight to the action class whose predicted score is larger than the actionness threshold. 
This module can be ineffective when the actionness thresholds are very large in the later stage of training. 
To investigate this, in Figure~\ref{fig:ablation_positive_weight_rate}, the \myBlue{blue curve} denotes the \myBlue{ratio} of scores exceeding the thresholds in all predictions, the \myGreen{green curve} denotes the \myGreen{training mAP} of each epoch with $10\times 1$ views. 
And we show how they evolve during training. 
We observe the \myBlue{ratio} is stable after training with AdaptFocus for $30$ epochs (i.e., the $70$-th epoch), while the performance of the model is still increasing. 
This phenomenon shows that AdaptFocus makes the trained models more confident in their predictions (i.e., the predicted scores become larger as the spike-actionness increases) during training, which demonstrates that our AdaptFocus framework is effective throughout the entire optimization process.
In Section~\ref{ablation_actioness_estimation} of \textbf{Appendix}, we provide more evidence to prove the reliability of the Spike-actioness Estimation.

\begin{figure}[!h]
\vspace{-0.4cm}
\setlength{\abovecaptionskip}{0cm}
\setlength{\belowcaptionskip}{0cm}
  \centering
   \includegraphics[width=0.8\columnwidth]{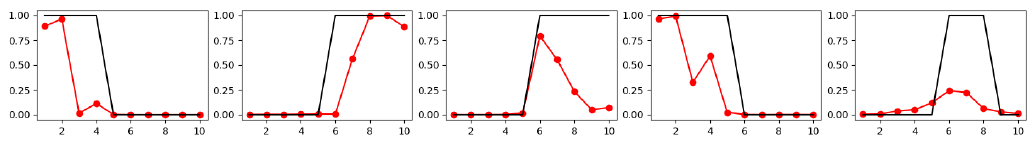}
   \caption{Visualization of the temporal distribution of the ground-truth (black curve) and prediction (red curve) of actions in a video. Each sub-figure shows an action. (Zoom in for the best view)}
   \label{fig:ablation_vis_predict_scores}
\vspace{-0.4cm}
\end{figure}

\noindent\textbf{- Qualitative Results.} For a long video that has six actions, Figure~\ref{fig:ablation_vis_predict_scores} comprises six sub-figures, where each sub-figure depicts the ground-truth (black curve) and prediction (red curve) of the corresponding action over temporal dimension (i.e., $10$ timestamps are evenly sampled from the long video). 
The visualization results show that AdaptFocus helps models learn to temporally localize actions even without such annotations during training. 
More visualizations are available in Section~\ref{more_visualizations} of the \textbf{Appendix}.

\subsection{A New Pipeline for Long-video Tasks}
\label{sec:LongVideoTasks}
Existing works in fine-grain long-video tasks~\cite{zheng2022weakly, lu2022set} (e.g., action segmentation) typically build their models on pre-extracted features of long videos, using action recognition models trained on short videos as extractors. 
However, the short-video data comes from significantly different domains, making the pre-extracted long-video features suffer from domain discrepancy. 
For the tasks that only allow access to weak annotations of long videos, our work provides a weakly supervised framework for end-to-end training recognition models as stronger extractors, without using additional annotations.
We evaluate our proposed weakly supervised feature extraction pipeline on three long-video tasks.

For each task, we evaluate different levels of supervision, including weakly supervised, semi-weakly supervised, and fully supervised settings. 
In each case, we re-evaluate the state-of-the-art model to demonstrate the effectiveness of our new feature extraction pipeline, denoted as \myPink{Weak-FEAT}. Since the choice of feature extraction method impacts task performance, we also re-extracted features using Kinetics-pretrained models (denoted as \myBlue{Kinc-FEAT}) and fully supervised models (denoted as \myGray{Full-FEAT}). 
However, it's worth noting that the fully supervised \myGray{Full-FEAT} features are often inaccessible in most scenarios.
For example, for the weakly supervised setting where the costly fully supervised annotations are not provided, using the fully supervised \myGray{Full-FEAT} features would actually violate the principles of the weakly supervised setting.
Two recognition models, i.e., I3D~\cite{carreira2017quo} and MViT-B-24~\cite{fan2021multiscale}, trained under different levels of supervision are used as extractors.

\subsubsection{Temporal Sentence Grounding}
\label{sec:tsg}

The temporal sentence grounding (TSG)~\cite{zheng2022weakly, li2023d3g, wang2022negative} aims to precisely localize the temporal boundaries of actions that are described by sentences. We use four evaluation metrics for this task (see details in~\cite{li2023d3g}). We show the results for the metric R1@IoU=0.5. The full results are presented in Section~\ref{full_result_tsg} of the \textbf{Appendix}. 

Table~\ref{tab:tsg_partial} shows our evaluation of three state-of-the-art models on the Charades dataset under different supervision: the weakly supervised CPL~\cite{zheng2022weakly}, semi-weakly supervised D3G~\cite{li2023d3g}, and fully supervised MMN~\cite{wang2022negative}. These models were originally developed using features extracted by Kinetics-pretrained models (i.e., \myBlue{Kinc-FEAT}). 
The evaluation with our \myPink{Weak-FEAT} pipeline reveals significant performance improvements for these methods. 
Moreover, across different extractors and supervision levels, models developed with our weakly supervised \myPink{Weak-FEAT} generally achieve comparable performance to those developed with the fully supervised \myGray{Full-FEAT}, which are inaccessible under weak and semi-weak supervised settings.

\subsubsection{Complex Activity Recognition}
\label{sec:ltar}

The Complex Activity Recognition (CAR)~\cite{hussein2019timeception} aims to recognize all actions within long videos by modeling long-term relations among actions, based on pre-extracted long-video features. 
The mean average precision metric (mAP) is used for evaluation. More information can be found in~\cite{zhou2023twinformer}. 

Existing CAR models are developed on the \myBlue{Kinc-FEAT} features that involve domain discrepancy. To show the effectiveness of our \myPink{Weak-FEAT} features, we evaluate one baseline, i.e., MLP (two linear-layers), and two state-of-the-arts, i.e., Timeception~\cite{hussein2019timeception} and GHRM~\cite{zhou2021graph}, using their official codebases. As shown in Table~\ref{tab:lar_breakfast}, all models exhibit significant improvements when using our \myPink{Weak-FEAT} features compared to \myBlue{Kinc-FEAT} features. 
Also, the performance gap between our \myPink{Weak-FEAT} and \myGray{Full-FEAT} features is small, indicating the effectiveness of our new pipeline even with much fewer annotations.

\subsubsection{Temporal Action Segmentation}
\label{sec:tas}

The temporal action segmentation (TAS)~\cite{bahrami2023much, lu2022set} task aims to segment video sequences into multiple non-overlapping action segments.
In this task, existing methods generally use both I3D-based RGB \myBlue{Kinc-FEAT} features and optical-flow \myBlue{Kinc-FEAT} features. 
To this end, we reform our RGB \myPink{Weak-FEAT} features by combining them with existing optical-flow \myBlue{Kinc-FEAT} features. 
For the weakly supervised POC model~\cite{lu2022set}, we use two metrics and report the average and maximum results over three runs. For the fully supervised LTContext model~\cite{bahrami2023much}, we employ five metrics, with F1 scores being the most important to evaluate the model. Please see details in their original papers~\cite{bahrami2023much, lu2022set}.
The results on Breakfast dataset in Table~\ref{tab:tas_breakfast} show that our weakly supervised \myPink{Weak-FEAT} features significantly boost model's performance compared to the currently used \myBlue{Kinc-FEAT} features.


\section{Conclusion}
\label{sec:conclusion}

This work proposes the first weakly supervised framework for end-to-end training action recognition models on long videos.
Experimental results on three datasets show that our framework can help weakly supervised models achieve comparable performance to fully supervised models. 
Moreover, with our framework, a weakly supervised feature extraction pipeline for long videos is proposed, which significantly improves the performance of existing state-of-the-arts on weakly supervised long-video tasks, providing better foundations for the development of long-video action understanding.

{\small
\bibliographystyle{unsrt}
\bibliography{neurips_2024}
}

\newpage

\appendix

\renewcommand\thefigure{{S\arabic{figure}}}
\renewcommand\thetable{{S\arabic{table}}}
\renewcommand\thesection{{S\arabic{section}}}

\setcounter{figure}{0} 
\setcounter{table}{0} 
\setcounter{section}{0} 

 \hrule height 4pt
\begin{center}
\Large\textbf{\myPink{Appendix} \\ Towards Weakly Supervised End-to-end Learning \\ for Long-Video Action Recognition}
\end{center}
\hrule height 1pt
\vskip 0.30in%

\begin{quote}
\textit{\textbf{THIS} work introduces the first weakly supervised end-to-end framework for training action recognition models on long-video data, balancing both efficiency and the cost of annotation during training. Leveraging the proposed framework, a new long-video feature extraction pipeline is established, which significantly improves the video representation for a wide range of long-video action understanding tasks. This is a long-term maintenance project, and will support more long-video datasets and long-video action understanding tasks in the future.}
\end{quote}




\section{More Experimental Analysis}
\label{supp:exps}

\subsection{Ablation on Hyper-parameters $\alpha$, $\beta$, $\theta$}
\label{ablation_parameter}

In our AdaptFocus framework, three hyper-parameters, $\alpha$, $\beta$, and $\theta$, are involved. They are empirically set to $5.0$, $3.0$, and $0.75$, respectively, for all experiments across different datasets and models. We conduct ablation studies by varying their values, keeping the other two parameters fixed when adjusting one.

\begin{figure}[h]
\centering
\begin{minipage}{0.39\linewidth}
\centering
   \includegraphics[width=0.7\linewidth]{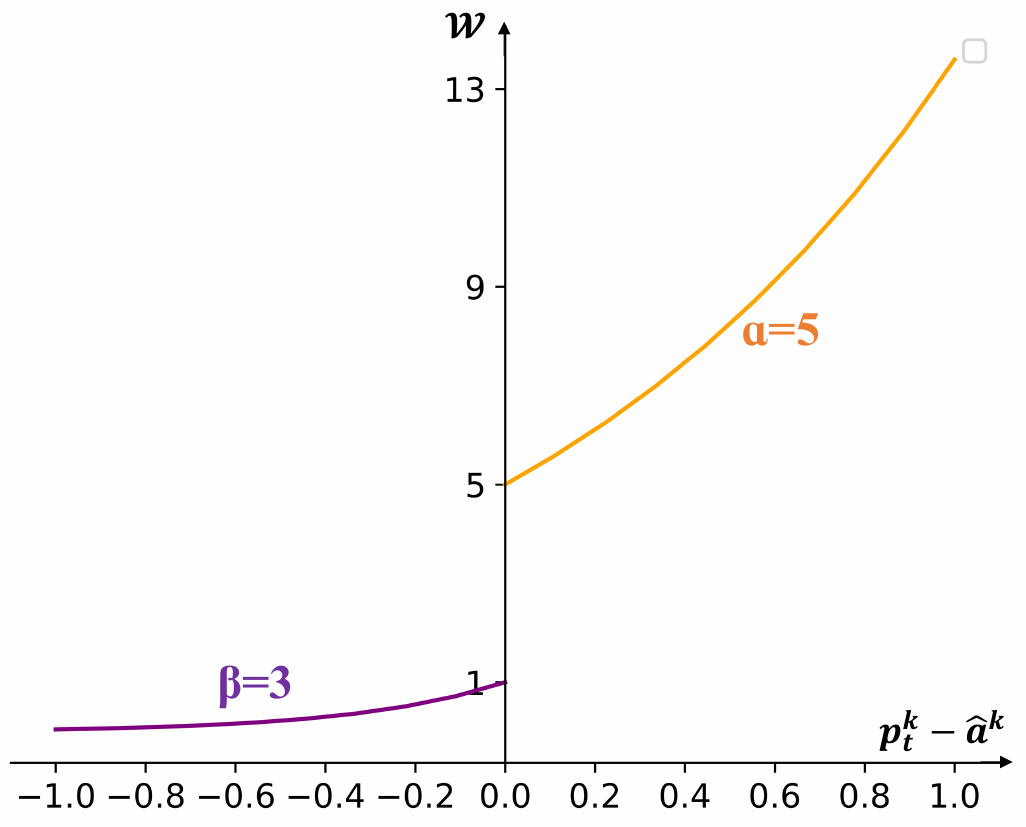}
   \caption{The curve of weighting function $\mathcal{W}$, where the x-axis represents the values of $p_t^k-\hat{a}^k$, and the y-axis represents the values of the function $\mathcal{W}$.}
   \label{fig:ablation_weight_func_sm}
\end{minipage}
\hfill
\begin{minipage}{0.59\linewidth}
\centering
\includegraphics[width=1.0\linewidth]{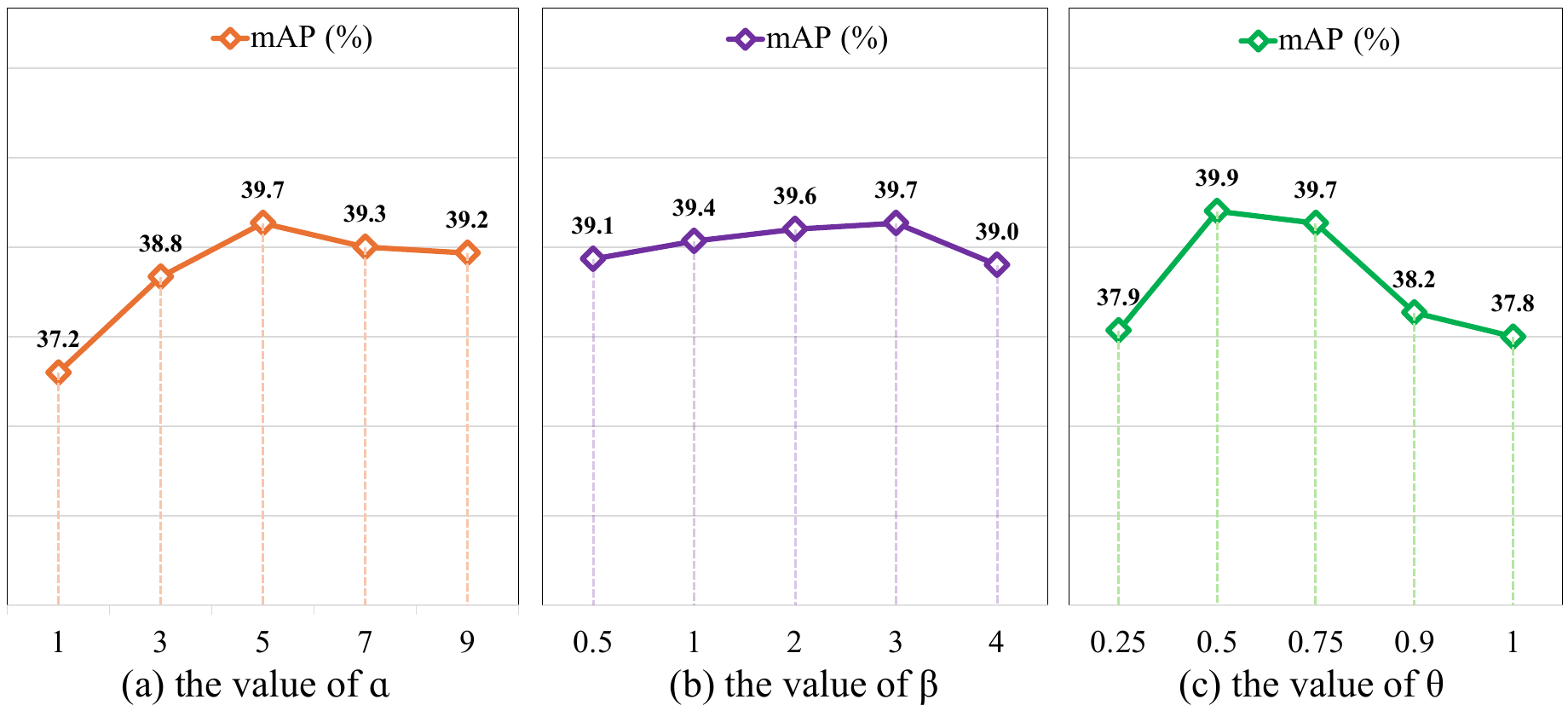}
   \caption{Ablation studies on the parameters $\alpha$, $\beta$, and $\theta$. They are empirically set to $5.0$, $3.0$, and $0.75$, respectively, for all experiments across different datasets and models.}
   \label{fig:ablation_hyper_params_sm}
\end{minipage}
\end{figure}

The parameters $\alpha>=1.0$ and $\beta>0$ modulate the scaling of weights as defined by the weighting function $\mathcal{W}$ in Eq.~\ref{eq:weight_func}, i.e., 
\begin{equation*}
\large
\mathcal{W}(p_t^k,\hat{a}^k) = 
\begin{cases}\alpha\cdot e^{(p_t^k-\hat{a}^k)}, & p_t^k>=\hat{a}^k, \\
e^{-\beta \cdot {(\hat{a}^k-p_t^k)}}, & p_t^k<\hat{a}^k. \\
\end{cases}
\end{equation*}
Figure~\ref{fig:ablation_weight_func_sm} illustrates the curve of $\mathcal{W}$, where $\alpha$ scales the positive part (i.e., $p_t^k-\hat{a}^k>=0$ and $\mathcal{W}>=1.0$) and $\beta$ scales the negative part (i.e., $p_t^k-\hat{a}^k<0$ and $\mathcal{W}<1.0$). Figures~\ref{fig:ablation_hyper_params_sm} (a) and (b) display the model's performance with different values of $\alpha$ and $\beta$, indicating the weighting function's robustness since the variation of the model's performance is small.

The parameter $\theta\in [0,1]$ determines the actionness threshold $\hat{a}$, i.e., $\hat{a}=\theta\cdot a$, where $a$ is the spike-actionness of each action instance. Figure~\ref{fig:ablation_hyper_params_sm} (c) examines the impact of different values of $\theta$, indicating that $\theta$ in the range of $0.5\sim 0.75$ is appropriate.

It is worth noting that, for different datasets and different models, our AdaptFocus framework shows its superiority without adjusting the values of these three hyper-parameters, which demonstrates the generality of our AdaptFocus framework.

\begin{table*}[!h]
\small
\setlength{\abovecaptionskip}{0cm}
\setlength{\belowcaptionskip}{0cm}
\caption{Temporal Sentence Grounding on \textbf{Charades} dataset.}
\begin{center}
\resizebox{1.0\linewidth}{!}
{
\begin{tabular}{c|c|c|c|c|c|c|c|c|c|c|c|c|c|c}
\specialrule{0.9pt}{0pt}{0pt}
\multicolumn{1}{c|}{\multirow{2}{*}{\makecell[c]{Super-\\vision}}} & \multicolumn{1}{c|}{\multirow{2}{*}{Model}} & \multicolumn{1}{c|}{Extr-} & \multicolumn{3}{c|}{R1@IoU=0.5 (\%)} & \multicolumn{3}{c|}{R1@IoU=0.7 (\%)} & \multicolumn{3}{c|}{R5@IoU=0.5 (\%)} & \multicolumn{3}{c}{R5@IoU=0.7 (\%)} \\
\cline{4-15}
\multicolumn{1}{c|}{} & \multicolumn{1}{c|}{} & \multicolumn{1}{c|}{actor} 
& \myGray{Full-FEAT} & \myBlue{Kinc-FEAT}   & \myPink{Weak-FEAT} 
& \myGray{Full-FEAT} & \myBlue{Kinc-FEAT}   & \myPink{Weak-FEAT} 
& \myGray{Full-FEAT} & \myBlue{Kinc-FEAT}   & \myPink{Weak-FEAT} 
& \myGray{Full-FEAT} & \myBlue{Kinc-FEAT}   & \myPink{Weak-FEAT} \\
\specialrule{0.9pt}{0pt}{0pt}

\addlinespace[0.5ex]
\specialrule{0.5pt}{0pt}{0pt}
\multicolumn{1}{c|}{\multirow{2}{*}{Weak}} & \multicolumn{1}{c|}{\multirow{2}{*}{\makecell[c]{CPL\\ \cite{zheng2022weakly}}}} & I3D 
& \myGray{48.6} & 39.6 & \textbf{48.1 \myPink{(+8.5)}}
& \myGray{21.4} & 18.6 & \textbf{21.4 \myPink{(+2.8)}}
& \myGray{84.0} & 81.4 & \textbf{83.9 \myPink{(+2.5)}}
& \myGray{50.7} & 49.2 & \textbf{50.4 \myPink{(+1.2)}} \\
\cline{3-15}
\multicolumn{1}{c|}{} & \multicolumn{1}{c|}{} & MViT 
& \myGray{53.7} & 47.8 & \textbf{51.7 \myPink{(+3.9)}}
& \myGray{24.8} & 21.8 & \textbf{23.2 \myPink{(+1.4)}}
& \myGray{85.7} & 84.6 & \textbf{85.2 \myPink{(+0.8)}}
& \myGray{52.9} & 50.4 & \textbf{52.6 \myPink{(+2.2)}} \\
\specialrule{0.5pt}{0pt}{0pt}

\addlinespace[0.5ex]
\specialrule{0.5pt}{0pt}{0pt}
\multicolumn{1}{c|}{Semi-} & \multicolumn{1}{c|}{\multirow{2}{*}{\makecell[c]{D3G\\ \cite{li2023d3g}}}} & I3D 
& \myGray{45.5} & 41.7 & \textbf{44.9 \myPink{(+3.2)}} 
& \myGray{20.0} & 18.8 & \textbf{20.0 \myPink{(+1.2)}} 
& \myGray{81.0} & 78.2 & \textbf{79.3 \myPink{(+1.1)}} 
& \myGray{48.5} & 48.0 & \textbf{48.4 \myPink{(+0.4)}} \\
\cline{3-15}
\multicolumn{1}{c|}{weak} & \multicolumn{1}{c|}{} & MViT 
& \myGray{49.3} & 46.0 & \textbf{49.1 \myPink{(+3.1)}} 
& \myGray{20.7} & 20.2 & \textbf{20.6 \myPink{(+0.4)}}
& \myGray{85.5} & 83.1 & \textbf{85.1 \myPink{(+2.0)}}
& \myGray{52.3} & 50.2 & \textbf{52.2 \myPink{(+2.0)}} \\
\specialrule{0.5pt}{0pt}{0pt}

\addlinespace[0.5ex]
\specialrule{0.5pt}{0pt}{0pt}
\multicolumn{1}{c|}{\multirow{2}{*}{Full}} & \multicolumn{1}{c|}{\multirow{2}{*}{\makecell[c]{MMN\\ \cite{wang2022negative}}}} & I3D 
& \myGray{55.8} & 49.4 & \textbf{55.7 \myPink{(+6.3)}}
& \myGray{35.5} & 29.8 & \textbf{35.4 \myPink{(+5.6)}}
& \myGray{87.5} & 85.8 & \textbf{87.5 \myPink{(+1.7)}}
& \myGray{64.4} & 60.5 & \textbf{64.0 \myPink{(+3.5)}} \\
\cline{3-15}
\multicolumn{1}{c|}{} & \multicolumn{1}{c|}{} & MViT
& \myGray{62.1} & 55.2 & \textbf{61.4 \myPink{(+6.2)}}
& \myGray{39.7} & 32.2 & \textbf{38.6 \myPink{(+6.4)}}
& \myGray{89.8} & 88.3 & \textbf{89.4 \myPink{(+1.1)}}
& \myGray{67.4} & 62.7 & \textbf{66.4 \myPink{(+3.7)}} \\
\specialrule{0.5pt}{0pt}{0pt}
\end{tabular}
}
\end{center}
\label{tab:tsg_full}
\end{table*}

\begin{figure*}[!h]
  \centering
   \includegraphics[width=1.0\linewidth]{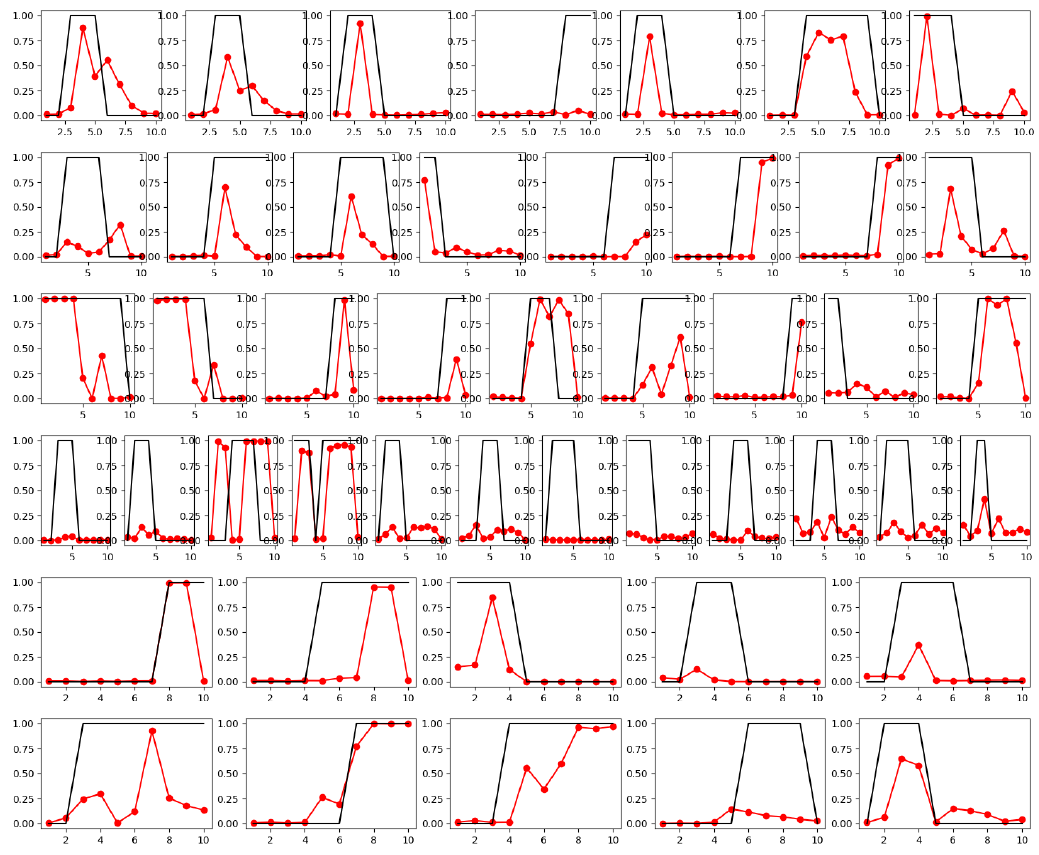}
   \caption{Visualization of the temporal distribution of the ground-truth (black curve) and prediction (red curve) of actions in videos. Each row presents a long video, where each sub-figure displays an action in the video. (Zoom in for the best view)}
   \label{fig:ablation_vis_predict_scores_sm}
\end{figure*}

\subsection{Full Results on Temporal Sentence Grounding}
\label{full_result_tsg}

Table~\ref{tab:tsg_full} presents full temporal sentence grounding results on the Charades dataset~\cite{sigurdsson2016hollywood}, evaluated using four metrics: R1@IoU=0.5, R1@IoU=0.7, R5@IoU=0.5, and R5@IoU=0.7.  Across different levels of supervision, model architectures, feature extractors, and evaluation metrics, performance on our \myPink{Weak-FEAT} features significantly outperforms that on the previously used \myBlue{Kinc-FEAT} features.


\subsection{More Visualizations}
\label{more_visualizations}

Similar to the visualization presented in Figure~\ref{fig:ablation_vis_predict_scores}, in Figure~\ref{fig:ablation_vis_predict_scores_sm}, we show additional results. Each row corresponds to a specific long video, where each sub-figure shows the temporal distribution of prediction (red curve) and groundtruth (black curve) of an action. The results show that our AdaptFocus framework successfully estimates the temporal location of each action in long videos, demonstrating the efficacy of the adaptive focus mechanism.

\subsection{Reliability of Spike-actionness Estimation}
\label{ablation_actioness_estimation}

The Action Saliency Estimation module is designed to accurately estimate the spike-actionness, i.e., the highest occurrence probability of each action instance in videos. This estimation is crucial for the effectiveness of the adaptive focus mechanism in the Action/Clip Focusing modules. 
In this work, the spike-actionness of each action is determined based on the classification scores from the learned action recognition model itself. 
Initially, the action recognition model is trained under noisy weak supervision during the warm-up phase, and then further trained through our AdaptFocus framework.

For each action instance in a video, predictions are obtained by applying the learned action recognition model to all video clips. We define the top-N success ratio as the probability that the temporal positions of the top-N highest predictions for each action instance across all clips fall within the temporal boundary of the action.
In Figure~\ref{fig:ablation_spike_estimated_accuracy_sm}, the evolution of the top-N success ratio during training is depicted. 
The red curve illustrates that even though the action recognition model is trained with noisy supervision (i.e., without AdaptFocus) for the initial $40$ epochs, the top-$1$ (i.e., spike-actionness estimation) success ratio achieves approximately $70\%$. This indicates that the model after warm-up training is an effective spike-actionness estimator. 
Moreover, under continued training with our AdaptFocus framework, this probability further increases, highlighting the framework's effectiveness.

Additionally, despite the temporal positions of the estimated spike-actionness of some action instances do not fall within their temporal boundaries (i.e., all clips within the action's temporal boundaries do not achieve the highest prediction), the high success ratios of top-$2$ and top-$3$ predictions after warm-up training indicate that clips within the action's temporal boundaries still estimate high-actionness, verifying the reliability of the learned action recognition model for estimating the actionness of actions.

\subsection{Analysis of Different Weighting Functions}
\label{ablation_weighting_func}
Drawing inspiration from the concept of selecting cleaner supervision in self-paced learning~\cite{DBLP:conf/nips/KumarPK10, DBLP:conf/icml/BengioLCW09}, our work adaptively focuses on cleaner actions and clips by adjusting their weights in the loss functions. 
Within the AdaptFocus framework, we employ the exponential weighting function (see Eq. \myRed{5}) in the Action Focusing module, and the binary weighting function (see Eq. \myRed{7}) in the Clip Focusing module. 
For instance, in the Action Focusing module, if the models believe that certain actions are likely to occur in the clip (i.e., $p_t^k>=\hat{a}^k$), the weights of these actions in the loss function will be larger than $1.0$. 
Leveraging this insight, in addition to the exponential weighting function, other different weighting functions used in self-paced learning can also be applied in our AdaptFocus framework. 
To this end, we design three additional weighting functions, i.e., constant, linear, and logarithmic weighting functions, and investigate their effectiveness.

\begin{figure}[h]
\centering
\begin{minipage}{0.49\linewidth}
\centering
   \includegraphics[width=1.0\linewidth]{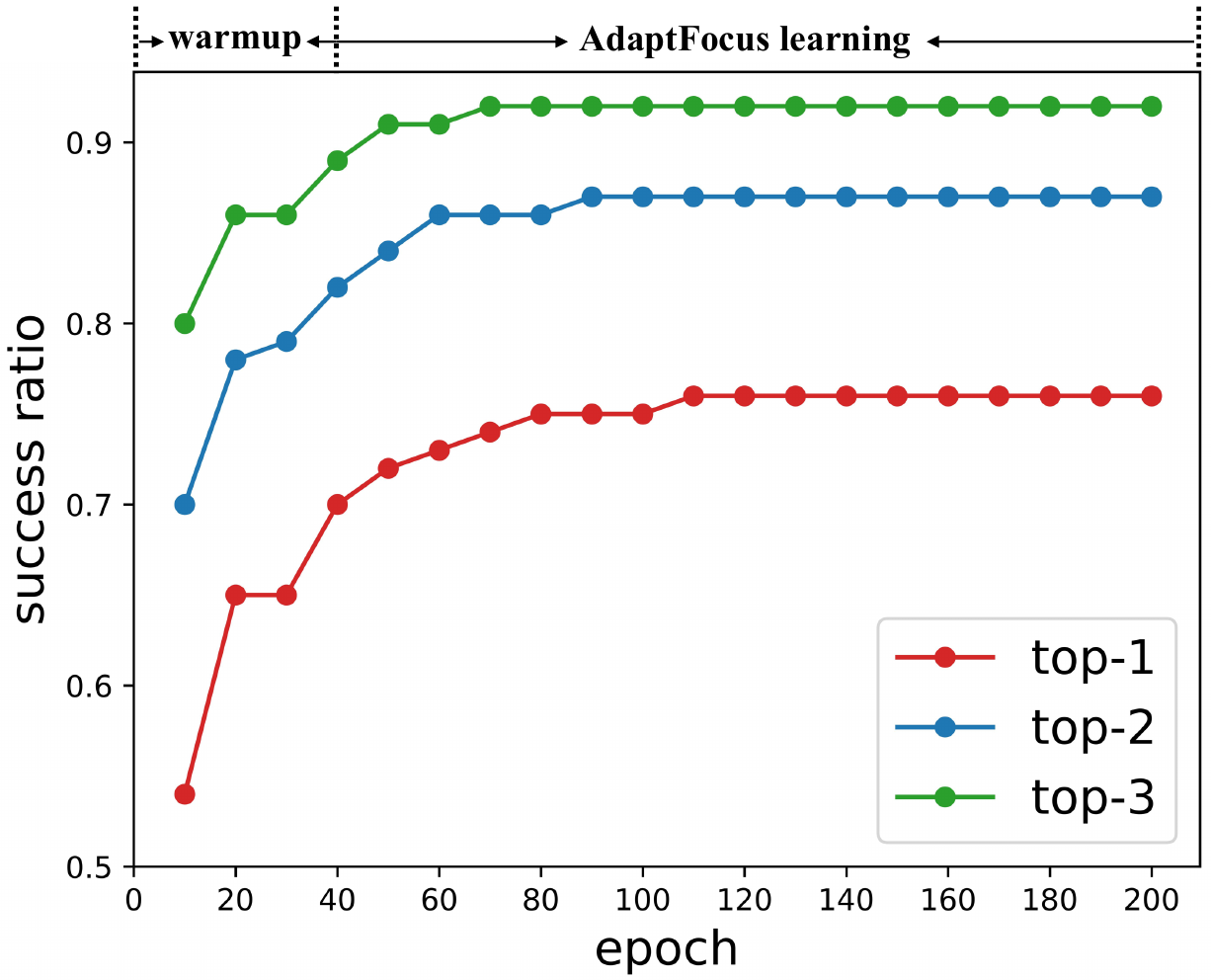}
   \caption{Illustration of the evolution of the top-N success ratio during training.}
\label{fig:ablation_spike_estimated_accuracy_sm}
\end{minipage}
\hfill
\begin{minipage}{0.50\linewidth}
\centering
\captionof{table}{Results of using different weighting functions in our AdaptFocus framework.}
\resizebox{0.8\textwidth}{!}
{
\begin{tabular}{c|c}
\specialrule{0.9pt}{0pt}{0pt}
Model & mAP (\%) \\
\hline
\myGray{Full (Clean)} & \myGray{40.0} \\
\hline
\myRed{Weak (Noisy)} & 34.1 \myRed{(-5.9)}\\
\hline
\addlinespace[0.5ex]
\hline
\multicolumn{2}{c}{\myPink{Weak (Ours)} } \\
\hline
Constant & 38.7 \myPink{(+4.6)} \\
Linear & 39.5 \myPink{(+5.4)} \\
Logarithmic & 39.2 \myPink{(+5.1)} \\
Exponential & \textbf{39.7 \myPink{(+5.6)}} \\
\specialrule{0.9pt}{0pt}{0pt}
\end{tabular}
}
\label{tab:ablation_different_weighting_functions}
\end{minipage}
\end{figure}

\noindent \textbf{- Constant weighting function} is defined as Eq.~\ref{eq:constant_weight_func}. In our AdaptFocus framework, we set $\alpha$ and $\beta$ to $5.0$ and $0.75$, respectively.

\begin{gather}
\large
\tag{S1}
\mathcal{W}(p_t^k,\hat{a}^k) = 
\begin{cases}\alpha, & p_t^k>=\hat{a}^k, \\
\beta, & p_t^k<\hat{a}^k, \\
\end{cases}
\label{eq:constant_weight_func}
\end{gather}

\noindent \textbf{- Linear weighting function} is defined as Eq.~\ref{eq:linear_weight_func}. The $\alpha$ and $\beta$ are set to $5.0$ and $1.0$, respectively. 

\begin{gather}
\large
\tag{S2}
\mathcal{W}(p_t^k,\hat{a}^k) = 
\begin{cases}\alpha\cdot [1+(p_t^k-\hat{a}^k)], & p_t^k>=\hat{a}^k, \\
\beta \cdot [1-(\hat{a}^k-p_t^k)], & p_t^k<\hat{a}^k, \\
\end{cases}
\label{eq:linear_weight_func}
\end{gather}

\noindent \textbf{- Logarithmic weighting function} is defined as Eq.~\ref{eq:logarithmic_weight_func}. We set $\alpha$ and $\beta$ to $5.0$ and $1.0$, respectively.

\begin{gather}
\large
\tag{S3}
\mathcal{W}(p_t^k,\hat{a}^k) = 
\begin{cases}\alpha\cdot \log_{e}[e + (p_t^k-\hat{a}^k)], & p_t^k>=\hat{a}^k, \\
\beta \cdot \log_{e}[ e - (\hat{a}^k-p_t^k)], & p_t^k<\hat{a}^k, \\
\end{cases}
\label{eq:logarithmic_weight_func}
\end{gather}

In Table~\ref{tab:ablation_different_weighting_functions}, we show the performance of our AdaptFocus framework when utilizing different weighting functions. The results show that for each weighting function, the model with AdaptFocus (i.e., \myPink{Weak (Ours)}) significantly outperforms the baseline model (i.e., \myRed{Weak (Noisy)}), effectively narrowing the performance gap with the oracle model (i.e., \myGray{Full (Clean)}). These findings underscore the effectiveness and robustness of the adaptive focus mechanism within the AdaptFocus framework.


\end{document}